\documentclass{article}

\usepackage[accepted]{icml2026}

\usepackage{microtype}
\usepackage{graphicx}
\usepackage{booktabs}
\usepackage{longtable}
\usepackage{placeins}
\usepackage{amsmath, amssymb}
\usepackage{hyperref}
\usepackage{url}
\usepackage{tikz}
\usetikzlibrary{arrows.meta, positioning, calc, backgrounds, fit}

\icmltitlerunning{A Causal DAG Prior for Synthetic Multivariate TSC}

\begin{document}

\twocolumn[
  \icmltitle{A Causal DAG Prior for Synthetic \\
             Time-Series Classification Datasets}

  \icmlsetsymbol{equal}{*}
  \begin{icmlauthorlist}
    \icmlauthor{Franco Martino O'Rourke}{mit}
    \icmlauthor{Ana Trisovic}{mit}
    \icmlauthor{Dimitris Bertsimas}{mit}
  \end{icmlauthorlist}
  \icmlaffiliation{mit}{Massachusetts Institute of Technology, Cambridge, MA, USA}
  \icmlcorrespondingauthor{Franco Martino O'Rourke}{franco03@mit.edu}

  \icmlkeywords{time-series classification, foundation models, in-context
                learning, synthetic data, causal DAG priors, TabPFN}

  \vskip 0.3in
]

\printAffiliationsAndNotice{}

\begin{abstract}
A Prior-data fitted Network learns the posterior predictive induced by 
its training prior; bringing this paradigm to multivariate 
time-series classification therefore calls for a synthetic generator 
that produces \emph{complete labelled datasets with temporal structure}.
We introduce a causal prior that synthesizes each dataset from a
randomly sampled DAG over typed nodes \emph{across two modalities}
(tabular attributes and time series), natively producing multivariate,
multi-class TSC datasets with cross-modal causal structure across
channels, timesteps and labels, a regime not addressed by existing
synthetic priors. To validate the
prior, we finetune TabPFN~v2.5 with minimal adaptations and evaluate on
75 UCR/UEA datasets within TabPFN's operating regime.  Finetuning on our generator
significantly outperforms both the unmodified upstream model and a
tabular-only ablation of the same prior (Wilcoxon signed-rank
$p=3.0\times 10^{-8}$ on ROC--AUC), isolating the contribution of the
cross-modal temporal structure.
\end{abstract}

\section{Introduction}
\label{sec:intro}

\paragraph{The two-stage status quo in TSC.}  Current foundation-model
pipelines for time-series classification
\citep{goswami2024moment,feofanov2025mantis} follow a \emph{two-stage}
recipe: a large transformer is pretrained on unlabelled time-series
corpora as an encoder of single series, and a dataset-specific
classifier (typically a linear head, a $k$-NN, or TabPFN on the
embeddings) is then fit on top of the frozen representations of the
training split.  Because the encoder ingests one sample at a time, it
cannot use the other training samples or their labels as context when
choosing which representation to extract, and the two stages optimise
different objectives.

\paragraph{In-context learning as an alternative.}  For tabular data,
TabPFN \citep{hollmann2023tabpfn,hollmann2025tabpfn2,tabpfn25}
demonstrates a single-stage alternative: one transformer is pretrained
on synthetic datasets sampled from a structural causal prior, and at
inference time predicts labels for a query set by conditioning on the
entire labelled training set in a single forward pass.  The same
paradigm, applied to TSC, would ingest
$(X,y_{\text{train}})\in\mathbb{R}^{n\times m\times T}\times
\{1,\dots,K\}^n$ as context and integrate temporal structure,
cross-channel interactions and label information from the start.  The
engine enabling this is a \emph{Bayesian prior-fitted network}
\citep{muller2022tcbi}: the model learns the posterior predictive
induced by a chosen data generator, so the synthetic prior directly
determines the inductive bias of the resulting classifier.

\paragraph{What is missing: a prior over complete labelled datasets.}
Real labelled TSC corpora could in principle train such a model, but
publicly available benchmarks are small and narrow in domain; and
existing synthetic priors for time series do not generate the object
we need.  KernelSynth~\citep{ansari2024chronos} and
CauKer~\citep{xie2025cauker} both sample time series from random
Gaussian-process kernels (composed via a structural causal model in
CauKer), but neither emits a class label, so neither produces a
complete labelled classification dataset.
Concurrently with our work, TiCT~\citep{yeh2025tict} pre-trains an
in-context classifier on synthetic data, but its prior is a binary
mixup between two univariate KernelSynth templates: not based on a
causal DAG, and not natively multivariate, multi-class or cross-modal
at the prior level.
What is absent is a prior that samples entire datasets
$(X,y)$ by modelling the causal relationships \emph{between two
modalities}, tabular attributes and time series.  In real systems it
is common to find simultaneously: tabular attributes that influence
temporal dynamics (\emph{e.g.}\ soil\,type $\to$ humidity$_t$); time
series that influence other time series across channels
(humidity$_t \to$ growth$_t$); variables with temporal self-causation
(humidity$_t \to$ humidity$_{t+1}$, humidity$_t \to$ growth$_{t+1}$);
and entire trajectories that determine a tabular outcome
(growth$_{1:T}\to$ yield).  Our prior samples a random causal graph
over typed nodes so that all of these relationships coexist, exposing
the model to the full space of cross-modal structures at training
time.

\paragraph{Contribution.}  Building on top of the tabular causal prior
of \citet{hollmann2025tabpfn2}, we introduce a synthetic causal prior
for multivariate TSC datasets that captures tabular, temporal,
cross-channel and label structure inside a single directed acyclic
graph (DAG).

To probe whether the prior contributes a useful learning signal we
finetune TabPFN~v2.5, applying only the minimal input/output edits
needed to make its tabular pipeline equivalent on time-series data
without destroying temporal structure (the goal is not to improve the
architecture but to isolate the contribution of the synthetic prior).

\section{Methodology}
\label{sec:method}

\subsection{Synthetic TSC data based on causal models}
\label{sec:method:generator}

We build on top of the causal-DAG prior of
\citet{hollmann2025tabpfn2}, extending it so that every sampled
dataset is a complete labelled multivariate TSC dataset.  Like the
tabular prior, our generator is based on structural causal models
(SCMs), which here make it possible to model causal relationships
\emph{between} two modalities (tabular attributes and time series)
inside a single graph.  Rather than unrolling a \emph{dynamic} SCM
recurrently over time~\citep{boeken2024dscm}, we favour a clean,
scalable design: an acyclic graph that injects temporal structure
non-recurrently, sidestepping the instability and scaling
difficulties that recurrent dynamics can exhibit under an untrained
prior~\citep{pascanu2013difficulty}.

\begin{figure}[t]
  \centering
  \includegraphics[width=\columnwidth]{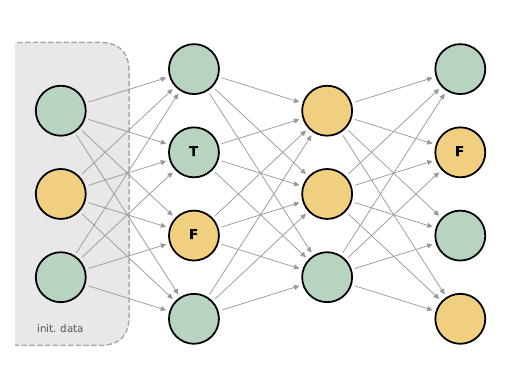}
  \caption{Each synthetic dataset is generated from a typed directed
  acyclic graph (DAG), where
  \textcolor[HTML]{3E7457}{\textbf{tabular}} nodes are scalars
  ($\in\mathbb{R}$) and
  \textcolor[HTML]{8A6010}{\textbf{series-valued}} nodes are
  trajectories ($\in\mathbb{R}^T$).  Root noise is drawn per sample
  from random scalar distributions
  (tabular) or from a Gaussian-process
  kernel composition fixed per dataset
  (series-valued).  Edges
  represent computational mechanisms including small neural networks,
  (causal) Conv1D, and discretisation.  Selected
  series-valued nodes (\textbf{F}) define
  the input $X$; a discretised tabular node
  (\textbf{T}) defines the class label $y$.  Optional
  post-processing is then applied.}
  \label{fig:dag}
\end{figure}

Our pipeline first samples dataset-level hyperparameters (e.g.\
dataset size, number of feature channels, series length, difficulty),
and then a DAG whose nodes are typed as either \emph{tabular} (a
scalar per sample) or \emph{series-valued} (a length-$T$ trajectory
per sample).  Edges can therefore carry causal effects both within
and across modalities (tabular $\to$ series, series $\to$ tabular).

To generate each sample, we propagate fresh initialisation data
through the DAG's root nodes.  Tabular roots are drawn from a random
normal or uniform distribution, as in \citet{hollmann2025tabpfn2}.
For each series root, a random kernel is sampled \emph{once per
dataset} (a composition of Linear/RBF/Periodic base kernels, in the
spirit of KernelSynth~\citep{ansari2024chronos}), and per-sample
trajectories are then drawn from the zero-mean Gaussian process it
defines; see Sec.~\ref{app:init}.

As these data traverse the edges of the computational graph we apply
a diverse set of computational mappings, chosen according to the type
of the child node.  At tabular child nodes the mapping is either a
small neural network with a pointwise nonlinearity or a discretisation
that turns a numerical output into a categorical one; at series child
nodes the mapping is a single random 1-D convolution with nonlinear
activation.  As an illustration, suppose a series node has four causal
parents, two tabular and two series-valued: each tabular parent is
replicated along the time axis to yield a constant length-$T$ channel,
the two series parents contribute one channel each, and the resulting
four-channel input is mapped to a length-$T$ output by a 1-D
convolution with causal padding, a randomly sampled dilation and a
nonlinear activation, encoding cross-channel mixing and temporal
dependence in a single operation.  As a second
illustration, suppose a tabular node has a single series-valued parent:
the parent trajectory is flattened to length~$T$ and the tabular node
output is a small neural network applied to this flat input, so that a
scalar outcome is determined by the entire trajectory of its parent.
At each edge we optionally add Gaussian noise; see
Sec.~\ref{app:edges} for details.

After traversing the causal graph, we read off a multivariate
trajectory $\mathbf{x}\in\mathbb{R}^{m\times T}$ at the sampled
series-valued feature nodes and a categorical target $y\in\{1,\dots,K\}$
at a tabular node whose mechanism is a discretisation.  Propagating
fresh initialisation data through the same graph $n$ times yields a
complete labelled TSC dataset
$(X,\mathbf{y})\in\mathbb{R}^{n\times m\times T}\times\{1,\dots,K\}^n$;
classification is therefore a readout on top of the SCM, not a
separate generator.

To further expose the model to common data challenges, we apply
predictive truncation (variable-length observations), value-level
missingness, random temporal granularity (strided pooling and
step-repeat) and per-channel value warps drawn from \{identity,
$\log$, $\exp$, squash, KDI, Kumaraswamy\}~\citep{hollmann2025tabpfn2};
see Sec.~\ref{app:post}.

\subsection{Using TabPFN as a probe of the prior}
\label{sec:method:arch}

To probe the learning signal of the prior we finetune the pretrained
TabPFN~v2.5 backbone~\citep{tabpfn25}.  We do not modify the backbone;
we apply only the minimal input/output edits needed to make TabPFN's
tabular pipeline equivalent on inputs of shape $n\times m\times T$
without destroying temporal structure.  The aim of these edits is not
to improve the architecture, but to keep it on equal footing so that
any change in performance reflects the synthetic prior alone.

If a TSC dataset is flattened as if it were tabular it becomes a
matrix $X\in\mathbb{R}^{n\times(m\cdot T)}$, with one column per
(channel, timestep) pair.  TabPFN~v2.5 forms input tokens by grouping
every $G=3$ consecutive columns (together with their missingness
indicators) and projecting the group through a linear tokeniser.
Our first adaptation is that the grouping is forced to respect the
channel boundary: we pad the time axis of each channel to the next
multiple of~$G$ before flattening, so that every token covers
$G$~consecutive timesteps of the \emph{same} channel.  The flattened
input thus becomes a sequence of short temporal \emph{patches} of
each channel, in the spirit of PatchTST~\citep{nie2023patchtst}.

TabPFN does not use a positional encoding; instead, the tokeniser
appends a fixed pseudo-random embedding to every column so the
attention layers can tell columns apart.  We leave this mechanism unchanged: after finetuning, when
the transformer attends to the entire labelled context of $n$ samples,
we expect it to be able to discover the correlation structure between patches
from the context itself even though the pseudo-random embeddings do not encode column order explicitly.  A second adaptation is normalisation: TabPFN's
default \texttt{normalize\_x} step $z$-scores each of the $G$
within-token positions independently across samples, which would wipe
out systematic temporal patterns inside a patch; we disable it and
instead apply $z$-scoring per channel before the tokeniser, so that
continuity within each patch is preserved.  The remaining preprocessing
is likewise computed per channel rather than per column.  Tokeniser
weights, transformer blocks and decoder are all inherited verbatim
from TabPFN~v2.5; we do not modify the ICL objective.  At inference we
ensemble $e\in\{1,8\}$ forward passes per dataset, with random channel
and class-index permutations between members; full schedule and
ensembling details are deferred to
Appendices ~\ref{app:train}--\ref{app:inference}.

\section{Experiments}
\label{sec:exp}

\subsection{Datasets}
\label{sec:exp:data}

We evaluate on the subset of UCR~\citep{dau2019ucr} and
UEA~\citep{bagnall2018uea} that fits within the operating regime of
TabPFN. TabPFN's native classification head is limited to $K \le 10$ classes \citep{hollmann2025tabpfn2}, and v2.5 applies per-estimator feature subsampling beyond 500 columns \citep{tabpfn25}; when a multivariate series is flattened to a tabular row, the latter limit applies to the product $m \cdot T$. We therefore retain all UCR and UEA datasets with $K \le 10$ and $m \cdot T \le 500$, obtaining 75 datasets. We use the default
train/test split supplied with each dataset
and report mean per-dataset accuracy and ROC--AUC; for multiclass
datasets we report macro-averaged one-vs-rest ROC--AUC.  Significance between
pairs of models is assessed with a Wilcoxon signed-rank test over the
75 datasets; global ranks are summarised with a critical-difference
(CD) diagram~\citep{demsar2006,middlehurst2024aeon}.

\subsection{Configurations}
\label{sec:exp:configs}

We compare four configurations, all built around the same TabPFN~v2.5
backbone, at two ensemble sizes $e\in\{1,8\}$:
\begin{description}
  \item[(A) Standard TabPFN.] The pretrained model is applied to the
        series flattened as a tabular row
        $X\in\mathbb{R}^{n\times(m\cdot T)}$, using the default TabPFN
        preprocessing and inference pipeline (column-wise
        \texttt{normalize\_x}, default tokeniser grouping,
        default SVD ensembling).  No modifications, no finetuning.
  \item[(B) + Our preprocessing.]  Same weights as (A), but with the
        adaptations of Sec.~\ref{sec:method:arch} (patch-aligned
        tokenisation, per-channel $z$-scoring,
        \texttt{normalize\_x} disabled).  Still no finetuning.
  \item[(C) + Finetuning (ours).]  (B) further finetuned on datasets
        drawn from our full generator (Sec.~\ref{sec:method:generator}).
  \item[(D) + Finetuning, tabular-only ablation.]  Same setup as (C),
        but every series-specific component of the generator is
        disabled: series-valued nodes, GP roots, Conv1D mechanisms,
        and the time-series augmentations of Sec.~\ref{app:post},
        reducing the prior to a purely tabular generator with $T=1$.
\end{description}

\subsection{Main result}

Table~\ref{tab:main} reports mean accuracy and AUC across the 75
benchmarks for the two ensemble sizes most commonly used in the TabPFN
literature.  Configuration (C) achieves the best accuracy and AUC at
both $e=1$ and $e=8$.  Our custom inference alone (B) \emph{loses} a
small amount of accuracy relative to (A), which is expected: our
inference preset trades away an SVD column-whitening step that helps
standard tabular data.  Finetuning on the full generator (C) more than
recovers the gap and overtakes the upstream model; finetuning on the
tabular-only ablation (D) does \emph{not}.

\begin{table}[t]
  \centering
  \small
  \caption{Mean accuracy and ROC--AUC on the 75 datasets.  Best per
  column in \textbf{bold}.  (C) uses our full prior; (D) is the
  tabular-only ablation defined in Sec.~\ref{sec:exp:configs}.}
  \label{tab:main}
  \setlength{\tabcolsep}{4pt}
  \begin{tabular}{@{}l@{\hskip 6pt}cccc@{}}
    \toprule
                               & \multicolumn{2}{c}{$e{=}1$}
                               & \multicolumn{2}{c}{$e{=}8$} \\
    \cmidrule(lr){2-3}\cmidrule(lr){4-5}
    Model                      & Acc.\ & AUC   & Acc.\ & AUC \\
    \midrule
    (A) Standard TabPFN        & .8436 & .9372 & .8525 & .9405 \\
    (B) Pretrained + our inf.  & .8378 & .9368 & .8489 & .9399 \\
    (D) Finetuned, tab.\ gen.  & .8366 & .9354 & .8493 & .9393 \\
    (C) Finetuned, \emph{ours} & \textbf{.8559} & \textbf{.9440}
                               & \textbf{.8575} & \textbf{.9444} \\
    \bottomrule
  \end{tabular}
\end{table}

\begin{figure}[t]
  \centering
  \includegraphics[width=\columnwidth]{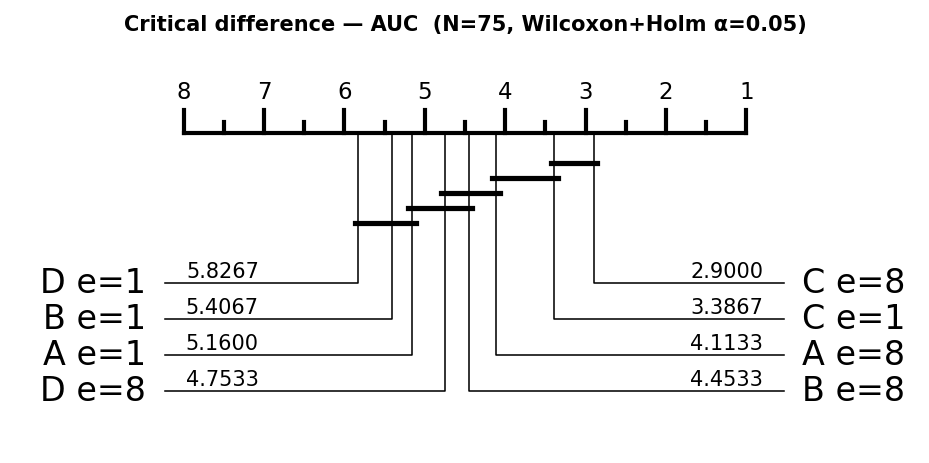}
  \caption{Critical-difference diagram (ROC--AUC) across all eight
  evaluated configurations, 75 datasets.  Lower rank is better.  The
  two variants of~(C) (ours) sit clearly ahead of (A), (B) and (D).}
  \label{fig:cd}
\end{figure}

Figure~\ref{fig:cd} visualises the same picture as a critical-difference
ranking over all eight configurations (four models $\times$ two ensemble
sizes): both variants of (C) sit clearly ahead of (A), (B) and (D).

\subsection{Ablation: are the series nodes necessary?}
\label{sec:ablation}

The central question is whether the \emph{series} mechanisms of the
generator (the GP-KernelSynth roots and the Conv1D interior nodes)
actually contribute signal, or whether finetuning on any DAG prior
that matches the input shape would suffice.
Table~\ref{tab:ablation} reports Wilcoxon signed-rank tests on both
mean ROC--AUC and mean accuracy across the 75 datasets.

\begin{table}[t]
  \centering
  \small
  \caption{Wilcoxon signed-rank tests across the 75 datasets, on
  both metrics.  ``Wins'' (computed on ROC--AUC) is the share of
  datasets on which model~A strictly beats model~B.}
  \label{tab:ablation}
  \setlength{\tabcolsep}{4pt}
  \begin{tabular}{lcccc}
    \toprule
    A vs B & Wins A & Wins B & $p_{\text{AUC}}$ & $p_{\text{Acc}}$ \\
    \midrule
    (C$_{e=1}$) vs (D$_{e=1}$) & 73.3\% & 12.0\% & $3.0\!\times\!10^{-8}$ & $1.1\!\times\!10^{-6}$ \\
    (C$_{e=8}$) vs (D$_{e=8}$) & 66.7\% & 14.7\% & $1.8\!\times\!10^{-7}$ & $8.3\!\times\!10^{-4}$ \\
    (C$_{e=1}$) vs (A$_{e=1}$) & 64.0\% & 22.7\% & $4.4\!\times\!10^{-4}$ & $4.1\!\times\!10^{-3}$ \\
    (C$_{e=1}$) vs (B$_{e=1}$) & 70.7\% & 14.7\% & $1.2\!\times\!10^{-6}$ & $1.5\!\times\!10^{-6}$ \\
    (D$_{e=1}$) vs (A$_{e=1}$) & 38.7\% & 48.0\% & $0.10$ & $0.20$ \\
    (D$_{e=1}$) vs (B$_{e=1}$) & 24.0\% & 57.3\% & $1.7\!\times\!10^{-4}$ & $0.43$ \\
    \bottomrule
  \end{tabular}
\end{table}

The results split into two consistent patterns across both metrics.
\emph{First}, the full-generator model (C) significantly outperforms
every other configuration: it beats the tabular ablation (D) at
$p_{\text{AUC}}{=}3.0\!\times\!10^{-8}$,
$p_{\text{Acc}}{=}1.1\!\times\!10^{-6}$, and the effect is preserved
at $e{=}8$ ($p_{\text{AUC}}{=}1.8\!\times\!10^{-7}$,
$p_{\text{Acc}}{=}8.3\!\times\!10^{-4}$), so the advantage is stable
across ensemble sizes.  \emph{Second}, the tabular ablation (D) gives
no positive signal over the baselines: it is statistically
indistinguishable from the upstream model (A) on both metrics
($p_{\text{AUC}}{=}0.10$, $p_{\text{Acc}}{=}0.20$), and on AUC is even
outperformed by the same backbone \emph{without} finetuning~(B)
($p_{\text{AUC}}{=}1.7\!\times\!10^{-4}$).  Finetuning on a
tabular-only DAG prior thus does not help, and can even hurt relative to
simply applying our patch-aligned inference.  The gain therefore
cannot be attributed to the finetuning recipe or the input/output
adaptations alone; it is the series mechanisms of the prior that
supply the signal.

\section{Conclusion}
\label{sec:conclusion}

We presented a causal prior for synthetic labelled TSC datasets that
types nodes in two modalities (tabular and series-valued) and
models causal effects \emph{between} them inside a single structural
causal model.  Using TabPFN~v2.5 as a probe, with only the minimal input/output edits
required to preserve temporal structure, we
showed that finetuning on this prior
improves over both the unmodified upstream model and a tabular-only
ablation of the same generator at the $3\!\times\!10^{-8}$ level on
ROC--AUC, isolating the contribution of the cross-modal temporal
structure.

\paragraph{Extension to richer structured data.}  Because the prior
operates at the level of \emph{node types}, the construction is not
specific to time series.  A \texttt{spatial} node type whose mechanism
is a 2-D convolution would yield image datasets; a
\texttt{spatio-temporal} type using a 3-D convolution would yield
video-like tensors; and mixing node types along the same DAG would
yield truly multimodal datasets, where the input contains both
tabular attributes and time series and the label depends on a
combination of both.  To our knowledge, no existing synthetic prior
models causal relationships \emph{between} modalities of this kind.

\paragraph{Limitations.}  This is a paper about the \emph{generator}:
we did not redesign the backbone, so the resulting probe is not a
state-of-the-art TSC foundation model and we do not benchmark against
full leaderboards; different patch sizes, a different tokeniser, or
pretraining from scratch on this prior are all likely to produce a
stronger model.  Evaluation is further restricted to the
$(m\cdot T\le 500,\,K\le 10)$ operating regime of TabPFN, which
excludes long or high-dimensional series, and the results only attest
to the incremental value of the series mechanisms over a tabular
prior, not to absolute TSC performance.  Finally, while the overall improvement supports the effectiveness of
causal DAGs that mix tabular and series-valued nodes, performance
degrades on some individual datasets, warranting further attention to
keep improving the prior coverage.

\newpage
\bibliographystyle{icml2026}
\bibliography{refs}

\newpage
\onecolumn
\appendix

\section{Details on the causal generative process}
\label{app:details}

This section gives the technical details of the generator of
Sec.~\ref{sec:method}.  An SCM $\mathcal{G}:=(\mathbf{Z},
\boldsymbol{\varepsilon})$ is a collection
$\mathbf{Z}:=(z_1,\dots,z_M)$ of structural assignments
$z_i = f_i(z_{\mathrm{PA}_{\mathcal{G}}(i)},\varepsilon_i)$, where
$\mathrm{PA}_{\mathcal{G}}(i)$ are the direct causes of $z_i$ in a
DAG~$\mathcal{G}$, $f_i$ is a deterministic mechanism, and
$\varepsilon_i$ is exogenous noise.  Every component below is defined
by a distribution; we use $\mathrm{U}$, $\mathrm{LogU}$ and
$\mathrm{LogU}_{\mathbb{Z}}$ for uniform, log-uniform and log-uniform
integer distributions.

\subsection{Graph structure sampling}
\label{app:dag}

The structural causal model underlying each dataset is based on a
DAG~$\mathcal{G}$.  We sample $\mathcal{G}$ as a layered feed-forward
graph: a root layer of width
$d_0\sim\mathrm{LogU}_{\mathbb{Z}}[a_d,b_d]$ is followed by
$L\sim\mathrm{LogU}_{\mathbb{Z}}[a_L,b_L]$ hidden layers of widths
$d_\ell\sim\mathrm{LogU}_{\mathbb{Z}}[a_w,b_w]$.  Every node in a
layer is initially connected to every node in the previous layer, and
edges are then dropped independently with probability
$p_{\mathrm{drop}}\sim\mathrm{U}[\alpha_{\mathrm{drop}},
\beta_{\mathrm{drop}}]$.  Disjoint subgraphs are allowed and lead to
features that are marginally independent of the target if they are
not connected to the target node, reflecting real-world scenarios
with uninformative predictors.

The node type $\tau_i\in\{\mathtt{tabular},\mathtt{series}\}$ is
sampled independently per node: with probability
$\pi_{\mathrm{S}}^{\mathrm{root}}$ at the root layer and
$\pi_{\mathrm{S}}^{\mathrm{hidden}}\sim\mathrm{U}[\alpha_{\mathrm{S}},
\beta_{\mathrm{S}}]$ at hidden layers the node is a \texttt{series}
node; otherwise it is \texttt{tabular}.  Three structural guarantees
are enforced by resampling: (i) $\mathcal{G}$ contains at least one
series root, at least one hidden series node and at least one tabular
node whose mechanism is a discretisation, so a classification target
exists; (ii) every non-root node has at least one parent; (iii) every
series node has at least one series parent, so series ancestry cannot
be broken by a tabular bottleneck.

\subsection{Computational edge mappings}
\label{app:edges}

In our implementation, each tabular SCM node and sample is represented
as a scalar in~$\mathbb{R}$, and each series SCM node and sample as a
vector in~$\mathbb{R}^T$.  When propagating data through the SCM, the
deterministic function $f_i$ at each non-root node maps the
concatenation of the parent outputs to its own output using one of
three types of computational modules, chosen according to the node
type:

\emph{1. Small neural networks (tabular nodes).}  We initialise a
weight vector $\mathbf{w}\in\mathbb{R}^{d_i}$ using Xavier
initialisation and apply the linear transformation
$\mathbf{w}^{\top}\mathbf{x}+b$ to the input vector~$\mathbf{x}$,
where $b\in\mathbb{R}$ is a bias.  After the linear projection, we
apply an element-wise nonlinear activation function
$\sigma:\mathbb{R}\to\mathbb{R}$, randomly sampled from a bank
including identity, logarithm, sigmoid, absolute value, sine,
hyperbolic tangent, squaring, power functions, smooth ReLU, step
function and modulo operation.  Tabular parents contribute a single
scalar to~$\mathbf{x}$; series parents are flattened along time and
contribute $T$ entries each.

\emph{2. Categorical feature discretisation (tabular nodes).}  To
generate categorical features from the numerical vectors at a node,
we map the input vector to the index of the nearest neighbour in a
set of $K$ randomly sampled prototype vectors
$\{\mathbf{p}_1,\dots,\mathbf{p}_K\}$ for a feature with $K$
categories,
$\kappa_i = \arg\min_{c}\|\mathbf{x}-\mathbf{p}_c\|$.  The discrete
index $\kappa_i$ will be observed in the feature set as a categorical
feature, or read off as the label if this node is selected as the
target (Sec.~\ref{app:target}).  To further use these discrete class
assignments in the computational graph they are re-embedded as
continuous scalars: we sample a per-class embedding
$v_c\sim\mathcal{N}(0,1)$ once per node and propagate
$z_i=v_{\kappa_i}$ to descendants.  The number of categories $K$ is
drawn from $\mathrm{LogU}_{\mathbb{Z}}[a_K,b_K]$.

\emph{3. One-dimensional convolutions (series nodes).}  To encode
cross-channel and temporal dependencies, series nodes apply a single
Conv1D.  Let $\mathbf{x}\in\mathbb{R}^{c_{\mathrm{in}}\times
T}$ be the concatenation of the parent channels along the channel
axis: scalar tabular parents are first broadcast to a constant
length-$T$ channel, and series parents contribute one channel each.
The series node output is
\begin{equation}
  z_i(t) \;=\; \sigma\!\left(
    \sum_{c=1}^{c_{\mathrm{in}}}\sum_{\ell=1}^{K}
    W_{c,\ell}\,\mathbf{x}_c\!\left(t-D(\ell-1)\right) + b
  \right),
\end{equation}
where the kernel length $K$ is drawn uniformly from a small set of
odd lengths plus the unit kernel, the dilation $D=\lfloor
2^{u}\rfloor$ with $u\sim\mathrm{U}[0,\log_2\tfrac{T-1}{K-1}]$, the
weights $W_{c,\ell}\sim\mathcal{N}(0,1)$ are mean-centred, and the
bias $b\sim\mathrm{U}[-\gamma,\gamma]$.  A dataset-level padding mode
(causal or centred) is drawn once per dataset and shared by all
series nodes, giving the generator a consistent notion of temporal
directionality.  The activation $\sigma$ is drawn from the same bank
as for tabular nodes, with the identity up-weighted so that a
recognisable temporal signal is preserved through deeper graphs.

\emph{4. Noise injection.}  At each non-root node we optionally add
Gaussian noise drawn from $\mathcal{N}(0,\sigma_{\varepsilon}^2
\mathbf{I})$ with probability $p_{\varepsilon}$.

\subsection{Initialization data sampling}
\label{app:init}

For each to-be-generated sample, we randomly generate initialisation
data $\boldsymbol{\varepsilon}$ that is inserted at the DAG root
nodes and then propagated through the computational graph.  The noise
is sampled according to the type of each root node:

\emph{1. Tabular roots.}  $\varepsilon\sim\mathcal{N}(0,
\sigma_{\varepsilon}^2)$ or $\varepsilon\sim\mathrm{U}[-a,a]$, one
family drawn per root.  If a tabular root is additionally assigned
the discretisation mechanism, its scalar is first mapped to a
categorical index and re-embedded as a scalar before being
propagated.

\emph{2. Series roots.}  The kernel $k$ is sampled \emph{once per
dataset} as a random composition of
$J\sim\mathrm{LogU}_{\mathbb{Z}}[a_J,b_J]$ base kernels drawn uniformly
from a bank $\{k_{\mathrm{Lin}},k_{\mathrm{RBF}},k_{\mathrm{Per}}\}$;
the composition is built by iteratively picking two kernels from the
current pool and combining them with either a sum or a product, until
a single kernel remains.  For each sample, the entire length-$T$
trajectory is then drawn from a Gaussian process,
\begin{equation}
  \mathbf{f}\sim\mathcal{GP}\!\left(0,\,k(t,t')\right),\qquad
  t\in\left\{\tfrac{0}{T-1},\dots,\tfrac{T-1}{T-1}\right\}.
\end{equation}
The kernel bank and compositional operators follow
KernelSynth~\citep{ansari2024chronos}.

\subsection{Post-processing / augmentation}
\label{app:post}

Each dataset is post-processed randomly with one or more of the
following operations:

\emph{1. Per-channel value warps.}  For some datasets, we apply a
per-channel transform drawn independently from $\{$identity, log,
exp, squash, KDI, Kumaraswamy$\}$, introducing nonlinear distortions
and scale differences.  The Kumaraswamy warp uses the
Kumaraswamy\,CDF as in~\citet{hollmann2025tabpfn2}.

\emph{2. Temporal granularity.}  A single transform is drawn per
dataset from $\{$identity, strided pooling, step-repeat$\}$.  Strided
pooling lowers the sampling rate by averaging over consecutive
timesteps; step-repeat artificially lowers the rate by emitting each
value several times.  This exposes the model to datasets measured at
different sampling rates of the same underlying process.

\emph{3. Predictive truncation.}  We designate a fraction of samples
as truncated: their observed series is cut to a random sub-horizon
$t_{\mathrm{obs}}<T$ while the target is still read from the full
trajectory of length~$T$.  This trains the model to produce
predictions from incomplete observations.

\emph{4. Value-level missing completely at random.}  To introduce
scenarios for dynamic imputation, a fraction $\rho_{\mathrm{miss}}$
of observations in each series is masked as missing, independently of
the data values.

Every sampled dataset is constrained to the PFN-eligibility region
$K\le K_{\max}$, $T\le T_{\max}$, $m\le m_{\max}$,
$m\cdot T\le (mT)_{\max}$, $n_{\mathrm{train}}\le n_{\max}$, matching
the regime of our probe.

\subsection{Target generation}
\label{app:target}

To generate the classification target, we select uniformly at random
one tabular node whose mechanism is a discretisation and set
$y_s=\kappa_{\star,s}$.  The number of classes is therefore bounded by
the number of prototypes at that node; we cap it at 10 classes.

\section{Training details}
\label{app:train}

We finetune a pretrained TabPFN~v2.5 backbone~\citep{tabpfn25} on
datasets drawn from the generator of App.~\ref{app:details}.  The
architecture is kept essentially unchanged; three adaptations are
made at the input/output boundary so that an
$n{\times}m{\times}T$ dataset can be ingested.  First, after sampling
a dataset we reshape it as $X\in\mathbb{R}^{n\times(m\cdot T)}$ and
pad the time axis to the next multiple of the model's column-group
size~$G$.  TabPFN forms a token by grouping $G$ consecutive columns;
padding in this way guarantees that each token covers $G$ consecutive
timesteps of the \emph{same} channel and that channels never mix
inside a token. To preserve this temporal structure during inference-time
ensembling, we also replace the default column shuffling used by
TabPFN with \emph{channel shuffling}, i.e., we permute entire channels
instead of individual columns so that consecutive timesteps always
remain adjacent. Second, we apply per-channel $z$-scoring before the
transformer, and we disable the encoder's internal \texttt{normalize\_x} step, which would $z$-score each of the $G$ within-token positions across samples and destroy systematic temporal patterns.  Tokeniser, transformer blocks and
decoder are inherited verbatim.

The training loss is the cross-entropy between the targets of
held-out samples of a synthetic dataset and the model prediction: for
a test set $(X_{\mathrm{test}},y_{\mathrm{test}})=\mathcal{D}_{
\mathrm{test}}$ drawn from our prior and the associated training
context $\mathcal{D}_{\mathrm{train}}$,
$\mathcal{L}=\mathbb{E}[-\log q_{\theta}(y_{\mathrm{test}}\mid
X_{\mathrm{test}},\mathcal{D}_{\mathrm{train}})]$~\citep{muller2022tcbi}.
We use AdamW with linear warmup followed by cosine
annealing.

\section{Inference details}
\label{app:inference}

At inference we feed a full labelled training set
$(X_{\mathrm{train}},y_{\mathrm{train}})\in\mathbb{R}^{n\times m\times
T}\times\{1,\dots,K\}^n$ together with a query batch
$X_{\mathrm{test}}\in\mathbb{R}^{n_{\mathrm{test}}\times m\times T}$
as one context, and the model returns a posterior over
$y_{\mathrm{test}}$ in a single forward pass.  As our model is not
fully permutation invariant, for each ensemble member we shuffle the
feature-channel order and, for classification, we additionally
permute the class labels, approximating order invariance.  We
ensemble $e\in\{1,8\}$ forward passes per dataset, averaging
predictions across members. We do not apply TabPFN's SVD
column-whitening preset at inference, which is tuned for tabular
columns rather than for the temporal columns of a single channel. Instead, we concatenate a pooled representation of each channel,
providing global channel-level context while preserving the original
temporal structure of the series.

\section{Generation cost}
\label{app:gencost}

On a single CPU core, generating 100 datasets takes
approximately 28 s (mean 0.24 s per dataset, median 0.03 s, max 8.3 s).
The high variance stems from the log-uniform distributions over $n\in[30,1400]$
and $T\in[6,2000]$: cost scales with
$n\cdot T$, so outliers arise when both are drawn large.


\section{Per-dataset results}
\label{app:per-dataset}

\begingroup
\scriptsize
\setlength{\tabcolsep}{3pt}
\begin{longtable}{lcccccccc}
\caption{Per-dataset ROC--AUC of the eight evaluated configurations on the 75 PFN-eligible UCR/UEA datasets.  Winner per dataset in \textbf{bold}.}\label{tab:per-dataset-auc}\\
\toprule
\textbf{Dataset} & \textbf{A $e{=}1$} & \textbf{A $e{=}8$} & \textbf{B $e{=}1$} & \textbf{B $e{=}8$} & \textbf{C $e{=}1$} & \textbf{C $e{=}8$} & \textbf{D $e{=}1$} & \textbf{D $e{=}8$} \\
\midrule
\endfirsthead
\multicolumn{9}{l}{\emph{(continued)}}\\
\toprule
\textbf{Dataset} & \textbf{A $e{=}1$} & \textbf{A $e{=}8$} & \textbf{B $e{=}1$} & \textbf{B $e{=}8$} & \textbf{C $e{=}1$} & \textbf{C $e{=}8$} & \textbf{D $e{=}1$} & \textbf{D $e{=}8$} \\
\midrule
\endhead
\midrule
\multicolumn{9}{r}{\emph{continued on next page}}\\
\endfoot
\bottomrule
\endlastfoot
AllGestureWiimoteX & .896 & .925 & .874 & .922 & .930 & \textbf{.937} & .864 & .920 \\
AllGestureWiimoteY & .912 & .940 & .951 & .952 & .954 & \textbf{.954} & .950 & .952 \\
AllGestureWiimoteZ & .913 & \textbf{.919} & .888 & .902 & .918 & .918 & .883 & .901 \\
ArrowHead & .909 & .910 & .935 & .927 & \textbf{.944} & .936 & .932 & .924 \\
BME & \textbf{1.000} & \textbf{1.000} & \textbf{1.000} & \textbf{1.000} & \textbf{1.000} & \textbf{1.000} & \textbf{1.000} & \textbf{1.000} \\
Beef & .981 & \textbf{.986} & .968 & .979 & .976 & .985 & .968 & .979 \\
CBF & .994 & .995 & .995 & .995 & \textbf{.998} & .997 & .995 & .995 \\
Chinatown & .996 & .995 & .996 & .996 & .994 & .996 & .996 & \textbf{.996} \\
ChlorineConcentration & .996 & \textbf{.999} & .991 & .994 & .996 & .996 & .991 & .994 \\
Coffee & \textbf{1.000} & \textbf{1.000} & \textbf{1.000} & \textbf{1.000} & \textbf{1.000} & \textbf{1.000} & \textbf{1.000} & \textbf{1.000} \\
DiatomSizeReduction & 1.000 & \textbf{1.000} & 1.000 & 1.000 & 1.000 & 1.000 & .999 & .999 \\
DistalPhalanxOutlineAgeGroup & .894 & .902 & .900 & .903 & .904 & \textbf{.905} & .900 & .903 \\
DistalPhalanxOutlineCorrect & .855 & .862 & \textbf{.884} & .878 & .877 & .878 & .880 & .877 \\
DistalPhalanxTW & .895 & \textbf{.909} & .885 & .890 & .890 & .895 & .883 & .889 \\
DodgerLoopDay & .903 & .909 & .821 & .897 & .885 & \textbf{.916} & .822 & .898 \\
DodgerLoopGame & .841 & .820 & \textbf{.904} & .868 & .891 & .875 & .903 & .873 \\
DodgerLoopWeekend & .986 & .987 & .989 & .990 & .990 & \textbf{.991} & .989 & \textbf{.991} \\
ECG200 & \textbf{.939} & .937 & .923 & .927 & .934 & .934 & .924 & .928 \\
ECG5000 & .948 & .950 & .946 & .947 & \textbf{.951} & .951 & .945 & .946 \\
ECGFiveDays & .981 & \textbf{.997} & .950 & .960 & .979 & .983 & .939 & .954 \\
ERing & .992 & .995 & .990 & .993 & .995 & \textbf{.996} & .990 & .993 \\
ElectricDevices & .891 & .902 & .892 & \textbf{.903} & .897 & .898 & .892 & .903 \\
FaceFour & \textbf{.992} & .991 & .990 & .990 & .986 & .987 & .989 & .990 \\
Fish & \textbf{.995} & .994 & .989 & .994 & .992 & .994 & .976 & .993 \\
FordA & .976 & \textbf{.977} & .925 & .943 & .944 & .951 & .924 & .944 \\
FordB & \textbf{.874} & .869 & .737 & .732 & .727 & .730 & .734 & .731 \\
FreezerRegularTrain & 1.000 & 1.000 & 1.000 & 1.000 & 1.000 & 1.000 & \textbf{1.000} & 1.000 \\
FreezerSmallTrain & .990 & .993 & .999 & .998 & \textbf{1.000} & 1.000 & .997 & .997 \\
GesturePebbleZ1 & .981 & .983 & .983 & .986 & \textbf{.986} & .985 & .983 & .985 \\
GesturePebbleZ2 & .954 & .954 & .946 & .959 & \textbf{.972} & .971 & .946 & .957 \\
GunPoint & .995 & .990 & .982 & .988 & .996 & \textbf{.997} & .981 & .989 \\
GunPointAgeSpan & \textbf{1.000} & \textbf{1.000} & 1.000 & \textbf{1.000} & \textbf{1.000} & \textbf{1.000} & \textbf{1.000} & \textbf{1.000} \\
GunPointMaleVersusFemale & \textbf{1.000} & \textbf{1.000} & \textbf{1.000} & \textbf{1.000} & \textbf{1.000} & \textbf{1.000} & \textbf{1.000} & \textbf{1.000} \\
GunPointOldVersusYoung & \textbf{1.000} & \textbf{1.000} & \textbf{1.000} & \textbf{1.000} & \textbf{1.000} & \textbf{1.000} & \textbf{1.000} & \textbf{1.000} \\
Ham & .834 & .816 & .816 & .814 & \textbf{.845} & .840 & .821 & .816 \\
ItalyPowerDemand & .994 & .994 & \textbf{.994} & .994 & .994 & .993 & .994 & .993 \\
JapaneseVowels & 1.000 & 1.000 & 1.000 & 1.000 & \textbf{1.000} & 1.000 & 1.000 & 1.000 \\
Lightning7 & .935 & .943 & .945 & .946 & .961 & \textbf{.962} & .942 & .945 \\
Meat & \textbf{1.000} & \textbf{1.000} & \textbf{1.000} & \textbf{1.000} & \textbf{1.000} & \textbf{1.000} & \textbf{1.000} & \textbf{1.000} \\
MedicalImages & .975 & .978 & .977 & .982 & .984 & \textbf{.984} & .977 & .982 \\
MelbournePedestrian & .999 & 1.000 & .999 & 1.000 & .999 & \textbf{1.000} & .999 & .999 \\
MiddlePhalanxOutlineAgeGroup & \textbf{.643} & .640 & .640 & .634 & .642 & .641 & .640 & .630 \\
MiddlePhalanxOutlineCorrect & .927 & .929 & .917 & .915 & .931 & \textbf{.932} & .912 & .913 \\
MiddlePhalanxTW & .789 & \textbf{.810} & .775 & .809 & .799 & .803 & .768 & .806 \\
MoteStrain & .959 & \textbf{.961} & .948 & .950 & .960 & .960 & .949 & .950 \\
OSULeaf & .905 & .911 & .896 & .895 & \textbf{.918} & .911 & .894 & .893 \\
PenDigits & 1.000 & 1.000 & 1.000 & 1.000 & \textbf{1.000} & 1.000 & 1.000 & 1.000 \\
PhalangesOutlinesCorrect & .919 & \textbf{.922} & .907 & .912 & .914 & .919 & .904 & .910 \\
PickupGestureWiimoteZ & .947 & .923 & .946 & .942 & \textbf{.953} & .950 & .944 & .944 \\
Plane & \textbf{1.000} & \textbf{1.000} & \textbf{1.000} & \textbf{1.000} & \textbf{1.000} & \textbf{1.000} & \textbf{1.000} & \textbf{1.000} \\
PowerCons & \textbf{1.000} & \textbf{1.000} & \textbf{1.000} & \textbf{1.000} & \textbf{1.000} & \textbf{1.000} & \textbf{1.000} & \textbf{1.000} \\
ProximalPhalanxOutlineAgeGroup & .939 & .941 & .936 & .941 & .941 & \textbf{.943} & .936 & .938 \\
ProximalPhalanxOutlineCorrect & .953 & \textbf{.966} & .960 & .955 & .954 & .961 & .958 & .954 \\
ProximalPhalanxTW & .943 & .936 & \textbf{.948} & .946 & .935 & .942 & .948 & .947 \\
RacketSports & .967 & .968 & .971 & .973 & .974 & \textbf{.974} & .972 & .973 \\
ShakeGestureWiimoteZ & .971 & .975 & .987 & .994 & .996 & \textbf{.998} & .984 & .994 \\
ShapeletSim & .459 & .468 & .468 & .466 & \textbf{.473} & .470 & .468 & .466 \\
SmoothSubspace & \textbf{1.000} & \textbf{1.000} & \textbf{1.000} & \textbf{1.000} & \textbf{1.000} & \textbf{1.000} & \textbf{1.000} & \textbf{1.000} \\
SonyAIBORobotSurface1 & .948 & \textbf{.959} & .933 & .939 & .958 & .958 & .935 & .941 \\
SonyAIBORobotSurface2 & .900 & .909 & .902 & .901 & \textbf{.912} & .911 & .904 & .903 \\
Strawberry & .995 & .995 & .995 & .994 & \textbf{.996} & .995 & .995 & .994 \\
Symbols & \textbf{.990} & .982 & .981 & .983 & .987 & .987 & .977 & .981 \\
SyntheticControl & 1.000 & 1.000 & \textbf{1.000} & \textbf{1.000} & 1.000 & \textbf{1.000} & \textbf{1.000} & \textbf{1.000} \\
ToeSegmentation1 & .602 & .612 & \textbf{.733} & .671 & .714 & .653 & .713 & .661 \\
ToeSegmentation2 & .759 & .795 & .945 & .954 & .949 & .953 & .945 & \textbf{.954} \\
Trace & .998 & .998 & \textbf{1.000} & \textbf{1.000} & \textbf{1.000} & \textbf{1.000} & \textbf{1.000} & \textbf{1.000} \\
TwoLeadECG & .971 & .988 & .975 & .971 & \textbf{.992} & .991 & .979 & .974 \\
TwoPatterns & 1.000 & \textbf{1.000} & 1.000 & 1.000 & 1.000 & \textbf{1.000} & 1.000 & 1.000 \\
UMD & \textbf{1.000} & \textbf{1.000} & \textbf{1.000} & \textbf{1.000} & \textbf{1.000} & \textbf{1.000} & \textbf{1.000} & \textbf{1.000} \\
UWaveGestureLibraryX & .968 & .970 & .968 & .970 & .970 & \textbf{.972} & .968 & .970 \\
UWaveGestureLibraryY & .951 & .954 & .951 & .955 & .951 & \textbf{.955} & .951 & .955 \\
UWaveGestureLibraryZ & .961 & \textbf{.963} & .957 & .962 & .961 & .962 & .957 & .963 \\
Wafer & 1.000 & 1.000 & 1.000 & 1.000 & \textbf{1.000} & 1.000 & 1.000 & 1.000 \\
Wine & .768 & .801 & .724 & .771 & .796 & \textbf{.809} & .719 & .760 \\
Yoga & \textbf{.945} & .945 & .931 & .935 & .935 & .944 & .930 & .935 \\
\midrule
\textit{Mean} & .937 & .941 & .937 & .940 & .944 & \textbf{.944} & .935 & .939 \\
\end{longtable}
\endgroup

\begingroup
\scriptsize
\setlength{\tabcolsep}{3pt}
\begin{longtable}{lcccccccc}
\caption{Per-dataset accuracy of the eight evaluated configurations on the 75 PFN-eligible UCR/UEA datasets.  Winner per dataset in \textbf{bold}.}\label{tab:per-dataset-acc}\\
\toprule
\textbf{Dataset} & \textbf{A $e{=}1$} & \textbf{A $e{=}8$} & \textbf{B $e{=}1$} & \textbf{B $e{=}8$} & \textbf{C $e{=}1$} & \textbf{C $e{=}8$} & \textbf{D $e{=}1$} & \textbf{D $e{=}8$} \\
\midrule
\endfirsthead
\multicolumn{9}{l}{\emph{(continued)}}\\
\toprule
\textbf{Dataset} & \textbf{A $e{=}1$} & \textbf{A $e{=}8$} & \textbf{B $e{=}1$} & \textbf{B $e{=}8$} & \textbf{C $e{=}1$} & \textbf{C $e{=}8$} & \textbf{D $e{=}1$} & \textbf{D $e{=}8$} \\
\midrule
\endhead
\midrule
\multicolumn{9}{r}{\emph{continued on next page}}\\
\endfoot
\bottomrule
\endlastfoot
AllGestureWiimoteX & .494 & .604 & .479 & .590 & .574 & \textbf{.631} & .460 & .597 \\
AllGestureWiimoteY & .583 & .664 & .636 & .694 & .667 & \textbf{.709} & .626 & .701 \\
AllGestureWiimoteZ & .554 & \textbf{.589} & .459 & .514 & .560 & .553 & .460 & .511 \\
ArrowHead & .703 & .726 & .766 & .743 & \textbf{.794} & .777 & .771 & .743 \\
BME & .993 & \textbf{1.000} & \textbf{1.000} & \textbf{1.000} & \textbf{1.000} & \textbf{1.000} & \textbf{1.000} & \textbf{1.000} \\
Beef & .867 & \textbf{.933} & .800 & .867 & .867 & .867 & .867 & .867 \\
CBF & .932 & .929 & .913 & .916 & .929 & \textbf{.937} & .913 & .914 \\
Chinatown & \textbf{.988} & .985 & .985 & .983 & .983 & .985 & .985 & .985 \\
ChlorineConcentration & .958 & \textbf{.973} & .952 & .959 & .965 & .968 & .953 & .958 \\
Coffee & \textbf{1.000} & \textbf{1.000} & \textbf{1.000} & \textbf{1.000} & \textbf{1.000} & \textbf{1.000} & \textbf{1.000} & \textbf{1.000} \\
DiatomSizeReduction & \textbf{.971} & \textbf{.971} & .964 & .967 & \textbf{.971} & \textbf{.971} & .964 & .967 \\
DistalPhalanxOutlineAgeGroup & .719 & .741 & .727 & .770 & .755 & .734 & .748 & \textbf{.784} \\
DistalPhalanxOutlineCorrect & .793 & .786 & .797 & \textbf{.801} & \textbf{.801} & \textbf{.801} & .790 & .797 \\
DistalPhalanxTW & .669 & \textbf{.691} & .655 & .647 & .655 & .662 & .640 & .655 \\
DodgerLoopDay & .623 & \textbf{.636} & .442 & .584 & .597 & \textbf{.636} & .442 & .584 \\
DodgerLoopGame & .748 & .748 & \textbf{.811} & .740 & .787 & .780 & .803 & .740 \\
DodgerLoopWeekend & \textbf{.984} & \textbf{.984} & \textbf{.984} & \textbf{.984} & \textbf{.984} & \textbf{.984} & \textbf{.984} & \textbf{.984} \\
ECG200 & \textbf{.880} & \textbf{.880} & .870 & .870 & \textbf{.880} & .870 & .860 & .870 \\
ECG5000 & .943 & .941 & .941 & .941 & \textbf{.944} & .944 & .939 & .938 \\
ECGFiveDays & .875 & \textbf{.967} & .863 & .872 & .906 & .923 & .848 & .862 \\
ERing & .896 & .919 & .881 & .911 & \textbf{.933} & .930 & .889 & .911 \\
ElectricDevices & .700 & .715 & .717 & \textbf{.740} & .727 & .737 & .718 & .739 \\
FaceFour & .909 & \textbf{.932} & .920 & \textbf{.932} & .920 & \textbf{.932} & .920 & \textbf{.932} \\
Fish & \textbf{.909} & .891 & .874 & .897 & .880 & \textbf{.909} & .783 & .903 \\
FordA & .913 & \textbf{.922} & .842 & .869 & .856 & .852 & .838 & .863 \\
FordB & .784 & \textbf{.786} & .659 & .600 & .604 & .585 & .658 & .600 \\
FreezerRegularTrain & .997 & .999 & \textbf{.999} & .999 & .998 & .998 & \textbf{.999} & .999 \\
FreezerSmallTrain & .932 & .936 & .949 & .951 & \textbf{.979} & .964 & .942 & .944 \\
GesturePebbleZ1 & .831 & .831 & .843 & .837 & \textbf{.849} & \textbf{.849} & .843 & .837 \\
GesturePebbleZ2 & .747 & .772 & .753 & \textbf{.785} & \textbf{.785} & .766 & .759 & \textbf{.785} \\
GunPoint & .953 & .940 & .913 & .913 & \textbf{.967} & \textbf{.967} & .913 & .927 \\
GunPointAgeSpan & .987 & \textbf{.997} & .994 & .994 & .994 & \textbf{.997} & .994 & .994 \\
GunPointMaleVersusFemale & \textbf{1.000} & \textbf{1.000} & .997 & .997 & \textbf{1.000} & \textbf{1.000} & .997 & .997 \\
GunPointOldVersusYoung & \textbf{1.000} & \textbf{1.000} & \textbf{1.000} & \textbf{1.000} & \textbf{1.000} & \textbf{1.000} & \textbf{1.000} & \textbf{1.000} \\
Ham & \textbf{.781} & .733 & .752 & .733 & .743 & .762 & .762 & .724 \\
ItalyPowerDemand & .973 & .972 & .971 & .973 & .965 & .968 & .970 & \textbf{.975} \\
JapaneseVowels & .986 & \textbf{.989} & .981 & .984 & .984 & .984 & .981 & .984 \\
Lightning7 & .699 & .671 & .658 & .740 & .712 & .740 & .644 & \textbf{.767} \\
Meat & \textbf{1.000} & \textbf{1.000} & .983 & .983 & .967 & .983 & .967 & .983 \\
MedicalImages & .803 & .800 & .843 & .838 & .847 & .846 & .849 & \textbf{.851} \\
MelbournePedestrian & .979 & .979 & .977 & .980 & .978 & \textbf{.981} & .977 & .980 \\
MiddlePhalanxOutlineAgeGroup & .584 & .617 & .552 & .578 & \textbf{.643} & .630 & .552 & .571 \\
MiddlePhalanxOutlineCorrect & .821 & \textbf{.849} & .828 & .842 & \textbf{.849} & .845 & .825 & .835 \\
MiddlePhalanxTW & .597 & .610 & .558 & .610 & \textbf{.630} & .623 & .571 & .623 \\
MoteStrain & .883 & \textbf{.884} & .871 & .871 & .879 & .883 & .872 & .871 \\
OSULeaf & .603 & .616 & .645 & .632 & \textbf{.661} & .653 & .645 & .636 \\
PenDigits & .977 & .981 & .985 & .984 & \textbf{.985} & .985 & .983 & .984 \\
PhalangesOutlinesCorrect & \textbf{.850} & .848 & .837 & .839 & .840 & .847 & .832 & .837 \\
PickupGestureWiimoteZ & \textbf{.760} & .700 & .740 & \textbf{.760} & .740 & .720 & .740 & .740 \\
Plane & \textbf{1.000} & .990 & \textbf{1.000} & .990 & .990 & .990 & \textbf{1.000} & .990 \\
PowerCons & .989 & \textbf{1.000} & .989 & .989 & \textbf{1.000} & \textbf{1.000} & .989 & \textbf{1.000} \\
ProximalPhalanxOutlineAgeGroup & .859 & .859 & .859 & .854 & .844 & .844 & \textbf{.863} & .854 \\
ProximalPhalanxOutlineCorrect & .897 & \textbf{.907} & .890 & \textbf{.907} & .904 & \textbf{.907} & .887 & .904 \\
ProximalPhalanxTW & .790 & .815 & .815 & .790 & \textbf{.820} & \textbf{.820} & \textbf{.820} & .785 \\
RacketSports & .842 & .875 & \textbf{.895} & .888 & .875 & .862 & \textbf{.895} & .888 \\
ShakeGestureWiimoteZ & .760 & .740 & .800 & \textbf{.900} & \textbf{.900} & .880 & .760 & \textbf{.900} \\
ShapeletSim & \textbf{.489} & .483 & .483 & .478 & \textbf{.489} & \textbf{.489} & .478 & .478 \\
SmoothSubspace & \textbf{1.000} & \textbf{1.000} & \textbf{1.000} & \textbf{1.000} & \textbf{1.000} & \textbf{1.000} & \textbf{1.000} & \textbf{1.000} \\
SonyAIBORobotSurface1 & .819 & \textbf{.822} & .750 & .772 & .809 & .812 & .747 & .764 \\
SonyAIBORobotSurface2 & .821 & \textbf{.834} & .824 & .827 & .832 & .831 & \textbf{.834} & .833 \\
Strawberry & \textbf{.981} & \textbf{.981} & .976 & .978 & \textbf{.981} & \textbf{.981} & .978 & .976 \\
Symbols & .896 & .890 & .865 & .893 & \textbf{.899} & \textbf{.899} & .862 & .885 \\
SyntheticControl & .990 & .987 & .993 & .997 & .993 & \textbf{1.000} & .993 & .997 \\
ToeSegmentation1 & .575 & .575 & .649 & .605 & \textbf{.662} & .636 & .645 & .601 \\
ToeSegmentation2 & .669 & .700 & .892 & .900 & \textbf{.908} & \textbf{.908} & .900 & .892 \\
Trace & .950 & .960 & \textbf{.990} & \textbf{.990} & \textbf{.990} & \textbf{.990} & \textbf{.990} & \textbf{.990} \\
TwoLeadECG & .904 & .944 & .903 & .895 & \textbf{.946} & .944 & .910 & .903 \\
TwoPatterns & .981 & \textbf{.998} & .984 & .997 & .987 & .998 & .984 & .997 \\
UMD & \textbf{1.000} & \textbf{1.000} & \textbf{1.000} & \textbf{1.000} & \textbf{1.000} & \textbf{1.000} & \textbf{1.000} & \textbf{1.000} \\
UWaveGestureLibraryX & .804 & .812 & .811 & \textbf{.814} & .811 & .812 & .810 & .814 \\
UWaveGestureLibraryY & .707 & .724 & .704 & \textbf{.724} & .707 & .723 & .702 & \textbf{.724} \\
UWaveGestureLibraryZ & .744 & \textbf{.763} & .727 & .757 & .751 & .760 & .727 & .757 \\
Wafer & .995 & .995 & .997 & .997 & .998 & .997 & .997 & \textbf{.998} \\
Wine & \textbf{.796} & .778 & .611 & .648 & .704 & .704 & .630 & .648 \\
Yoga & \textbf{.875} & .865 & .864 & .863 & .860 & .862 & .863 & .866 \\
\midrule
\textit{Mean} & .844 & .852 & .838 & .849 & .856 & \textbf{.858} & .837 & .849 \\
\end{longtable}
\endgroup

\section{Datasets excluded from the UCR/UEA archives}
\label{app:excluded}

\begingroup
\scriptsize
\setlength{\tabcolsep}{5pt}
\begin{longtable}{llccccl}
\caption{Datasets from the UCR and UEA archives excluded from our evaluation because they fall outside TabPFN's operating regime ($K{>}10$ and/or $m{\cdot}T{>}500$).  Totals: 83 excluded; 75 retained.}\label{tab:excluded}\\
\toprule
\textbf{Dataset} & \textbf{Archive} & $\mathbf{K}$ & $\mathbf{m}$ & $\mathbf{T}$ & $\mathbf{m{\cdot}T}$ & \textbf{Exclusion reason} \\
\midrule
\endfirsthead
\multicolumn{7}{l}{\emph{(continued)}}\\
\toprule
\textbf{Dataset} & \textbf{Archive} & $\mathbf{K}$ & $\mathbf{m}$ & $\mathbf{T}$ & $\mathbf{m{\cdot}T}$ & \textbf{Exclusion reason} \\
\midrule
\endhead
\midrule
\multicolumn{7}{r}{\emph{continued on next page}}\\
\endfoot
\bottomrule
\endlastfoot
ACSF1 & UCR & 10 & 1 & 1460 & 1460 & $m{\cdot}T{=}1460$ \\
Adiac & UCR & 37 & 1 & 176 & 176 & $K{=}37$ \\
BeetleFly & UCR & 2 & 1 & 512 & 512 & $m{\cdot}T{=}512$ \\
BirdChicken & UCR & 2 & 1 & 512 & 512 & $m{\cdot}T{=}512$ \\
Car & UCR & 4 & 1 & 577 & 577 & $m{\cdot}T{=}577$ \\
CinCECGTorso & UCR & 4 & 1 & 1639 & 1639 & $m{\cdot}T{=}1639$ \\
Computers & UCR & 2 & 1 & 720 & 720 & $m{\cdot}T{=}720$ \\
CricketX & UCR & 12 & 1 & 300 & 300 & $K{=}12$ \\
CricketY & UCR & 12 & 1 & 300 & 300 & $K{=}12$ \\
CricketZ & UCR & 12 & 1 & 300 & 300 & $K{=}12$ \\
Crop & UCR & 24 & 1 & 46 & 46 & $K{=}24$ \\
EOGHorizontalSignal & UCR & 12 & 1 & 1250 & 1250 & $K{=}12$, $m{\cdot}T{=}1250$ \\
EOGVerticalSignal & UCR & 12 & 1 & 1250 & 1250 & $K{=}12$, $m{\cdot}T{=}1250$ \\
Earthquakes & UCR & 2 & 1 & 512 & 512 & $m{\cdot}T{=}512$ \\
EthanolLevel & UCR & 4 & 1 & 1751 & 1751 & $m{\cdot}T{=}1751$ \\
FaceAll & UCR & 14 & 1 & 131 & 131 & $K{=}14$ \\
FacesUCR & UCR & 14 & 1 & 131 & 131 & $K{=}14$ \\
FiftyWords & UCR & 50 & 1 & 270 & 270 & $K{=}50$ \\
Fungi & UCR & 18 & 1 & 201 & 201 & $K{=}18$ \\
GestureMidAirD1 & UCR & 26 & 1 & 360 & 360 & $K{=}26$ \\
GestureMidAirD2 & UCR & 26 & 1 & 360 & 360 & $K{=}26$ \\
GestureMidAirD3 & UCR & 26 & 1 & 360 & 360 & $K{=}26$ \\
HandOutlines & UCR & 2 & 1 & 2709 & 2709 & $m{\cdot}T{=}2709$ \\
Haptics & UCR & 5 & 1 & 1092 & 1092 & $m{\cdot}T{=}1092$ \\
Herring & UCR & 2 & 1 & 512 & 512 & $m{\cdot}T{=}512$ \\
HouseTwenty & UCR & 2 & 1 & 2000 & 2000 & $m{\cdot}T{=}2000$ \\
InlineSkate & UCR & 7 & 1 & 1882 & 1882 & $m{\cdot}T{=}1882$ \\
InsectEPGRegularTrain & UCR & 3 & 1 & 601 & 601 & $m{\cdot}T{=}601$ \\
InsectEPGSmallTrain & UCR & 3 & 1 & 601 & 601 & $m{\cdot}T{=}601$ \\
InsectWingbeatSound & UCR & 11 & 1 & 256 & 256 & $K{=}11$ \\
LargeKitchenAppliances & UCR & 3 & 1 & 720 & 720 & $m{\cdot}T{=}720$ \\
Lightning2 & UCR & 2 & 1 & 637 & 637 & $m{\cdot}T{=}637$ \\
Mallat & UCR & 8 & 1 & 1024 & 1024 & $m{\cdot}T{=}1024$ \\
MixedShapesRegularTrain & UCR & 5 & 1 & 1024 & 1024 & $m{\cdot}T{=}1024$ \\
MixedShapesSmallTrain & UCR & 5 & 1 & 1024 & 1024 & $m{\cdot}T{=}1024$ \\
NonInvasiveFetalECGThorax1 & UCR & 42 & 1 & 750 & 750 & $K{=}42$, $m{\cdot}T{=}750$ \\
NonInvasiveFetalECGThorax2 & UCR & 42 & 1 & 750 & 750 & $K{=}42$, $m{\cdot}T{=}750$ \\
OliveOil & UCR & 4 & 1 & 570 & 570 & $m{\cdot}T{=}570$ \\
PLAID & UCR & 11 & 1 & 1345 & 1345 & $K{=}11$, $m{\cdot}T{=}1345$ \\
Phoneme & UCR & 39 & 1 & 1024 & 1024 & $K{=}39$, $m{\cdot}T{=}1024$ \\
PigAirwayPressure & UCR & 52 & 1 & 2000 & 2000 & $K{=}52$, $m{\cdot}T{=}2000$ \\
PigArtPressure & UCR & 52 & 1 & 2000 & 2000 & $K{=}52$, $m{\cdot}T{=}2000$ \\
PigCVP & UCR & 52 & 1 & 2000 & 2000 & $K{=}52$, $m{\cdot}T{=}2000$ \\
RefrigerationDevices & UCR & 3 & 1 & 720 & 720 & $m{\cdot}T{=}720$ \\
Rock & UCR & 4 & 1 & 2844 & 2844 & $m{\cdot}T{=}2844$ \\
ScreenType & UCR & 3 & 1 & 720 & 720 & $m{\cdot}T{=}720$ \\
SemgHandGenderCh2 & UCR & 2 & 1 & 1500 & 1500 & $m{\cdot}T{=}1500$ \\
SemgHandMovementCh2 & UCR & 6 & 1 & 1500 & 1500 & $m{\cdot}T{=}1500$ \\
SemgHandSubjectCh2 & UCR & 5 & 1 & 1500 & 1500 & $m{\cdot}T{=}1500$ \\
ShapesAll & UCR & 60 & 1 & 512 & 512 & $K{=}60$, $m{\cdot}T{=}512$ \\
SmallKitchenAppliances & UCR & 3 & 1 & 720 & 720 & $m{\cdot}T{=}720$ \\
StarLightCurves & UCR & 3 & 1 & 1024 & 1024 & $m{\cdot}T{=}1024$ \\
SwedishLeaf & UCR & 15 & 1 & 128 & 128 & $K{=}15$ \\
UWaveGestureLibraryAll & UCR & 8 & 1 & 945 & 945 & $m{\cdot}T{=}945$ \\
WordSynonyms & UCR & 25 & 1 & 270 & 270 & $K{=}25$ \\
Worms & UCR & 5 & 1 & 900 & 900 & $m{\cdot}T{=}900$ \\
WormsTwoClass & UCR & 2 & 1 & 900 & 900 & $m{\cdot}T{=}900$ \\
ArticularyWordRecognition & UEA & 25 & 9 & 144 & 1296 & $K{=}25$, $m{\cdot}T{=}1296$ \\
AtrialFibrillation & UEA & 3 & 2 & 640 & 1280 & $m{\cdot}T{=}1280$ \\
BasicMotions & UEA & 4 & 6 & 100 & 600 & $m{\cdot}T{=}600$ \\
CharacterTrajectories & UEA & 20 & 3 & 119 & 357 & $K{=}20$ \\
Cricket & UEA & 12 & 6 & 1197 & 7182 & $K{=}12$, $m{\cdot}T{=}7182$ \\
DuckDuckGeese & UEA & 5 & 1345 & 270 & 363150 & $m{\cdot}T{=}363150$ \\
EigenWorms & UEA & 5 & 6 & 17984 & 107904 & $m{\cdot}T{=}107904$ \\
Epilepsy & UEA & 4 & 3 & 206 & 618 & $m{\cdot}T{=}618$ \\
EthanolConcentration & UEA & 4 & 3 & 1751 & 5253 & $m{\cdot}T{=}5253$ \\
FaceDetection & UEA & 2 & 144 & 62 & 8928 & $m{\cdot}T{=}8928$ \\
FingerMovements & UEA & 2 & 28 & 50 & 1400 & $m{\cdot}T{=}1400$ \\
HandMovementDirection & UEA & 4 & 10 & 400 & 4000 & $m{\cdot}T{=}4000$ \\
Handwriting & UEA & 26 & 3 & 152 & 456 & $K{=}26$ \\
Heartbeat & UEA & 2 & 61 & 405 & 24705 & $m{\cdot}T{=}24705$ \\
InsectWingbeat & UEA & 10 & 200 & 20 & 4000 & $m{\cdot}T{=}4000$ \\
LSST & UEA & 14 & 6 & 36 & 216 & $K{=}14$ \\
Libras & UEA & 15 & 2 & 45 & 90 & $K{=}15$ \\
MotorImagery & UEA & 2 & 64 & 3000 & 192000 & $m{\cdot}T{=}192000$ \\
NATOPS & UEA & 6 & 24 & 51 & 1224 & $m{\cdot}T{=}1224$ \\
PEMS-SF & UEA & 7 & 963 & 144 & 138672 & $m{\cdot}T{=}138672$ \\
PhonemeSpectra & UEA & 39 & 11 & 217 & 2387 & $K{=}39$, $m{\cdot}T{=}2387$ \\
SelfRegulationSCP1 & UEA & 2 & 6 & 896 & 5376 & $m{\cdot}T{=}5376$ \\
SelfRegulationSCP2 & UEA & 2 & 7 & 1152 & 8064 & $m{\cdot}T{=}8064$ \\
SpokenArabicDigits & UEA & 10 & 13 & 65 & 845 & $m{\cdot}T{=}845$ \\
StandWalkJump & UEA & 3 & 4 & 2500 & 10000 & $m{\cdot}T{=}10000$ \\
UWaveGestureLibrary & UEA & 8 & 3 & 315 & 945 & $m{\cdot}T{=}945$ \\
\end{longtable}
\endgroup

\end{document}